# A Popperian Falsification of Artificial Intelligence - Lighthill Defended

Steven Meyer
smeyer@tdl.com
April 30, 2020

Abstract

The area of computation called artificial intelligence (AI) is falsified by describing a previous 1972 falsification of AI by British mathematical physicist James Lighthill. How Lighthill's arguments continue to apply to current AI is explained. It is argued that AI should use the Popperian scientific method in which it is the duty of scientists to attempt to falsify theories and if theories are falsified to replace or modify them. The paper describes the Popperian method and discusses Paul Nurse's application of the method to cell biology that also involves questions of mechanism and behavior. It is shown how Lighthill's falsifying arguments especially combinatorial explosion continue to apply to modern AI. Various skeptical arguments against the assumptions of AI mostly by physicists especially against Hilbert's philosophical programme that defined knowledge and truth as provable formal sentences. John von Neumann's arguments from natural complexity against neural networks and evolutionary algorithms are discussed. Next the game of chess is discussed to show how modern chess experts have reacted to computer chess programs. It is shown that currently chess masters can defeat any chess program using Kasperov's arguments from his 1997 Deep Blue match and aftermath. The game of 'go' and climate models are discussed to show computer applications where combinatorial explosion may not apply. The paper concludes by advocating studying computation as Peter Naur's Dataology.

## 1. Introduction

This paper applies the method of falsification discovered by Karl Popper to show that artificial intelligence (AI) programs are not intelligent and in fact are just normal computer application programs in which programmers express their ideas by writing computer code or are data analysis programs that produce output for use in human problem solving.

AI is meaningless metaphysics in the Popperian sense of metaphysics based on a number of incorrect assumptions and dogmas that was falsified by James Lighthill in his evaluation of AI for the British science funding agency (Lighthill[1972]). This paper defends Lighthill's 20th century falsification of AI and explains how it applies to current AI. One important part of Lighthill's argument provided explanations of why computation problems believed to require human intelligence were so combinatorially complex that no program could solve them.

This analysis of combinatorial explosion explains why some problems that seemingly require an exponential number of search steps such as the game 'go' or atom combination analysis for medicine development are really not so complex. The game 'go' does not run into the combinatorial explosion barrier because primitives and move type are so few. Searching for feasible molecules with medicinal properties does not run into combinatorial explosion because there are numerous chemical and physical constraints on possible realizable molecules (from quantum mechanics and energy conservation particularly). In contrast chess programs run into combinatorial explosion because move choice combinations are numerous and involve long term strategic planning (non specific and variably bound choices)..

Material is presented that the author developed from being encouraged to criticize AI as a 1960s Stanford University undergraduate and from a talk given to Paul Feyerabend's philosophy of science seminar while the author was a computer science (CS in Literature and Science School) student at UC Berkeley. In order to understand why Lighthill's criticism falsifies the AI research programme and why his arguments still apply to AI now in the third decade of the 21st century in spite of vast improvements in computer speed and capacity, it is necessary to understand the development of modern computers primarily by physicists after WWII. The paper uses recent historical scholarship to explain Lighthill's background assumptions and shows how that background knowledge also falsifies current AI.

It was not just Lighthill who was skeptical of AI. Physicists in general are critics of AI. See for example Roger Penrose's two books that show the impossibility of artificial minds (Penrose[1994] and Penrose[2016]). There is a file in David Bohm's archive at Birbeck College in London that appears to be Bohm planning to write a paper criticizing AI that I believe was never written.

The current lack of criticism of AI research programmes may be related to a historical accident of academic organization at Stanford University in the middle of the 20th century. The accident was that for some reason Stanford decided not to offer academic appointments to Stanford Linear Accelerator (SLAC) physicists who had academic appointments at the institution they came from. Assistant SLAC director Mathew Sands discussed this in his American Physical Society (APS) interview (Sands[1987], p. 192). Administrator Albert Bowker who was responsible for starting Stanford's computer science as an academic discipline also discusses the problem with SLAC appointments (Bowker[1979], p. 6). Physicists encouraged the study of computation. For example, Niklaus Wirth developed his various computer languages while working at and being funded by SLAC.

The effect of few physicists with academic appointments and I claim William Miller's incorrect understanding of the questionable intellectual standing of AI was that Stanford computer science became the same as the Stanford AI Lab. William Miller was the administrator responsible for Stanford computer science after Bowker left Stanford, For example, assistant CS professor Jeff Barth was fired in 1977 because he refused to work at the Stanford AI Lab. Miller explained the idea for Stanford computer science this way. "I think we tended to focus on fairly rigorous problems that could be recognized as rigorous problems. We followed the paradigms of more rigorous disciplines and established it as a science as opposed to an engineering or applied discipline" (Miller[1979], p. 11). Unfortunately, the result of the Stanford organization resulted in almost total suppression of criticism of AI.

## 2. What is Popperian falsification

Falsification is a method discovered by Karl Popper that argues general statements do not have scientific merit. Only singular statements that Popper calls basic statements have meaning. Basic statements (propositions) have simple structure. Such statements can be disproven either by scientific experiments or by logic (Popper[1968], p. 74). Popper's major contribution to the philosophy of science requires scientists to criticizes one's own theories to the fullest extent possible so that false theories can be modified or replaced. Popperians believe scientific method consists of numerous bold conjectures that are then tested and if falsified, eliminated or modified. Popper's method calls for bold conjecture followed by stringent criticism.

Popper's original falsification theory developed in the late 1920s and early 1930s is called naive falsification (Lakatos[1999], pp. 64-85). The theory was improved and generalized by Popper and his colleagues during most of the 20th century. I am using the term Popperian philosophy in a sense that includes the modifications and improvement to Popper's theory mostly carried out at the London School of Economics not just by Popper but also by: Imre Lakatos, Paul Feyerabend and Thomas Kuhn. The other aspects of Popperian methodology is most clearly expressed by Imre Lakatos as the *Methodology of Scientific Research Programmes (MSRP)* (Lakatos[1970]). There were disagreements among the Popperians about questions of emphasis but not about methodology or importance of rationality in science. James Lighthill, as holder of the Lucasian chair in applied mathematics at Cambridge University, was familiar with and part of the milieu that developed Popperian theory.

It is important to understand that falsification needs to be a necessary condition for scientific research. It is not sufficient because there are situation for which falsified theories need to be kept because for example there is no acceptable alternative. Lakatos calls this research programme competition (Lakatos[1970]).

Falsification as a theory in philosophy of science is usually discussed in terms of physics because the developers of falsification were trained as physicists. Physics is possibly not a good fit for study of AI methodology because in AI there is no mechanism or functional explanation involved in attempting to understand physical reality (describe fields or particle interactions for example). The connection to cell biology that attempts to understand and utilize the mechanisms of cell behavior is closer. Paul Nurse in his 2016 Popper Memorial Lecture discusses the importance of bold conjectures and diligent attempts to eliminate incorrect theory by falsification (Nurse[2016]). Nurse also discussed data analysis in cell biology. For readers unfamiliar with Popperian falsification, the Nurse lecture provides an excellent introduction.

Donald Gillies offers a different criticism of current machine learning style AI (Gillies[1996]). Gillies attributes the type of induction used in AI to early 17th century philosopher Sir Francis Bacon. "A scientist [robot] following the method should begin by making a large number of careful observations. Then, from the mass of data, laws should be extracted by a process known as *Induction*" (p. 2). Gillies shows that Popperian falsification is a better method than induction. However, Gillies defends expert system style AI. This paper shows that AI is just applications programming. Where Gillies sees expert system AI, I see applications programming.

AI has proven difficult to falsify. However, falsification was developed to falsify theories difficult to criticize. One important example is Popperian falsification of the scientific aspects of Freudian psychoanalysis. Freudian psychoanalysis was finally falsified by Frank Cioffi who successfully showed that it is pseudo science (Cioffi[1985]). AI is difficult to criticize in the same sense psychoanalysis is difficult to criticize because methods continually change often as a reaction to falsification attempts. Different aspects of AI are presented as encyclopedic lists in the face of criticism but are construed as correct theories of intelligence when not being criticized. See Nils Nilsson's *The Quest for Artificial Intelligence - A History of ideas and achievements* (Nilsson[2010]) for an example. There is circular drift of current popular AI methods with methods falling in and out of favor similar to the pattern exhibited by the scientific claims of Freudians.

Falsification of AI is important because it is claimed that computational intelligence is now so successful that discussions of ethical issues involving how inferior humans will deal with the superior intellect of AI robots are required. The activity I view as computer application programming is even sometimes called "AI supremacy" in the popular literature.

## 3. Lighthill's falsification of AI

Lighthill's falsification of AI is quite simple (Lighthill[1972]) and I claim continues to apply to AI in spite of changes mostly in vastly faster computers that execute machine instructions in parallel and new names for algorithms such as "deep learning" that replaces alpha beta heuristics to improve logic resolution algorithms to implement intelligence. Lighthill argues AI is just CS described using the language of human intelligence and views computers and computation as tools for expressing people's ideas.

Lighthill divides AI into three areas. Category A: Automation (feedback control engineering), Category C: computer based studies of the central nervous system, and Category B: the bridge area between A and B that is supposedly going to provide the magic synergy that allows creation of intelligent robots (p. 3). For example, current deep learning would fall into areas A and B. It falls into category B because it involves automatic logical deduction without any need for a person to program ideas into the algorithm, but also it is in category A because it "looks beyond conventional data processing to the problems involved in large-scale data banking and retrieval" (p. 5). I think Lighthill is arguing here that AI studies normal computer science but rephrases problems in terms of human attributes (p. 7 paragraph 2).

According to Lighthill for control engineering it should not matter how the engineering is accomplished. Lighthill writes in the section discussing category A: "Nevertheless it (AI) must be looked at as a natural extension of previous work of automation of human activities, and be judged by essentially the same criteria" (p. 4 paragraph 4). After more than 40 years of computer development, programmable digital computers are usually the best choice for control engineering. In modern terms current feedback control engineering is based on improvements in camera technology allowing more precise location measurements and more complex feedback. Advances and cost reductions in computer and storage technology allow large amounts of data to be processed faster and at lower cost.

In criticizing AI's approach to area C since obviously it makes sense to study neurophysiology, Lighthill distinguishes syntactic automation as advocated currently by AI versus conceptual automation (p.

6). He asks if "a device that mimics some human function somehow assists in studying and making a theory of the function of the central nervous system" (p. 6 paragraph 4).

Lighthill criticizes the use of mathematical logic in AI by arguing practical use runs into a combinatorial explosion (p. 10 paragraph 5) and argues there are difficulties in storing axioms favored by logicians versus heuristic knowledge favored by AI (p. 10, paragraph 6). In my view this is the crucial falsifier of AI. Namely, although Lighthill was attempting to provide a neutral assessment of AI, he did not believe in the Hilbert Programme that is the central tenet of AI.

Lighthill also discusses organization problems with AI methodology. He questions claims such as "robots better than humans by 2000" (p. 13) (now probably replace with 2030). Lighthill as an mathematical physicist also discusses the combinatorial explosion that allows humans to solve problems that can not be solved by formal algorithms.

## 3.1 Understanding Lighthill's falsification in modern terms

In 1972 Lighthill falsified AI by showing its individual claims were false and by arguing there was no unified subject but rather just normal problems in the area of computation involving computer applications and study of data. AI researchers were not convinced at the time, I think, because Lighthill did not make his Popperian view of science clear. The remainder of this paper discusses how 1970s scientific background knowledge particularly in the physics and applied mathematics areas falsifies current AI methods. The discussion is possible because of recent scholarship especially in the areas of Hilbert's philosophical programme and in the study of John von Neumann's thinking during the development of digital computers.

## 4. Skepticism toward Hilbert's programme of truth as formal proof

In the 1920s, mathematician David Hilbert conjectured that knowledge and truth consists solely of all sentences that can be proven from axioms. Hilbert's original conjecture was a mathematical problem. However, it was interpreted as a philosophical theory in which truth became formal proof from axioms. A paradigmatic example is the Birkhoff and Von Neumann formalization of quantum mechanics as axiomatized logic (Birkhoff[1936], Popper]1968] attempted to falsify it). Hilbert's programme as the basic assumption of AI is that knowledge about the world can be expresses as formal sentences. Knowledge is then expressed as formulas that can be derived using logic (usually predicate calculus) from other sentences about the world that are true.

In addition to the belief that knowledge is formal sentences, the foundation of AI is the belief that the Church'-Turing Thesis (Copeland[2015]) is true. Namely, that nothing can exist outside of formally proven sentences. proven from axioms. In the AI community this dogma is beyond criticism. However, the philosophical Hilbert programme was abandoned starting in the 1930s for various reasons. The reason most often given is that Godel's incompleteness results showed the Hilbert programme could not succeed. The Hilbert programme is still believed in the logic area and AI seems to be grasping at straws by attempting to mitigate the Godel disproof by finding in practice areas where Godel's results do not apply. Zach[2015] Stanford Encyclopedia of Philosophy article discusses some attempts to mitigate Godel's results. See Detlefsen[2017] for a more skeptical view of Hilbert's programme.

There were a number of other reasons Hilbert's philosophical programme was rejected as discussed below. These other reasons explain why the AI argument that since people have intelligence, computer programs can also have intelligence is incorrect. In the view of AI, the problem is just building faster computers and developing better algorithms so that computers can discover and learn the formal sentences in people's heads.

## 4.1 Von Neumann's argument - automata and neural networks useless at high levels of complexity

During the second half of the 20th century, John von Neumann's work on computers and computations was widely accepted. Publication of Von Neumann's work on computing did not occur until years after Lighthill's falsification was written (in particular Aspray[1990], Neumann[2005] and Kohler[2001]). Lighthill was certainly familiar with Von Neumann's work.

John Von Neumann studied automata and neural networks when he was developing his Von Neumann computer architecture. Neumann expressed his skepticism toward linguistics and automata as sources of AI algorithms in discussing problems with formal neural networks when he wrote:

*The insight that a formal neuron network can do anything which you can describe in words a very important insight and simplifies matters enormously at low complication levels. It is by no means certain that it is a simplification on high complication levels. It is perfectly possible that on high complication levels the value of the theorem is in the reverse direction, namely, that you can express logics in terms of these efforts and the converse may not be true (Von Neumann[1966], quoted in Aspray[1990], note 94, p. 321).*

Von Neumann also considered and rejected current AI methodology when he developed the Von Neumann computer architecture. In a 1946 paper with Herman Goldstine on the design of a digital computer Von Neumann wrote that some sort of intuition had to be built into programs instead of using brute force searching (Aspray[1990], p. 62). Edward Kohler (Kohler[2000]), p. 118) describes von Neumann's discovery in developing modern computer architecture in an article "Why von Neumann Rejected Carnap's Duality of Information Concepts" as:

*Most readers are tempted to regard the claim as trivial that automata can simulate arbitrarily complex behavior, assuming it is described exactly enough. But in fact, describing behavior exactly in the first place constitutes genuine scientific creativity. It is just such a prima facie superficial task which von Neumann achieved in his [1945] famous explication of the "von Neumann machine" regarded as the standard architecture for most post World-War-II computers.*

The problem context in the area of operations research solution space searching that influenced both von Neumann and Lighthill was pre computer algorithmic operations research (see Budiansky[2013] for the detailed story). Understanding the limitations of combinatorial explosion arises naturally from that experience.

### 4.2 Skepticism toward linguistics and formal languages in computing

Starting with Ludwig Wittgenstein in the late 1930s, skepticism toward linguistics and especially formal languages become prevalent. Wittgenstein's claim was that mathematical (and other) language was nothing more than pointing (Wittgenstein[1930]). The Popperians and English science in general were receptive to Wittgenstein and his "pointing" philosophy of mathematics. Popperians avoided linguistic philosophy because they viewed it as creating more problems than it solved. I read Lighthill's falsification as assuming this attitude toward language. Modern AI still claims knowledge and truth is limited to provable formal sentences.

### 4.3 Physicist skepticism towards mathematics as axiomatized logic

In my view there was a more important reason for the rejection of Hilbert's programme. Physicists were always skeptical toward axiomatized mathematics. Albert Einstein in his 1921 lecture on geometry expresses this skepticism. Einstein believed that formal mathematics was incomplete and disconnected from physical reality. Einstein stated:

*This view of axioms, advocated by modern axiomatics, purges mathematics of all extraneous elements. ... such an expurgated exposition of mathematics makes it also evident that mathematics as such cannot predicate anything about objects of our intuition or real objects (Einstein[1921]).*

Niels Bohr argued that first comes the conceptual theory then the calculation. John von Neumann expressed the physicist attitude with a story relating a conversation with founder of quantum physics Wolfgang Pauli. "If a mathematical proof is what matters in physics, you would be a great physicist" (Thirring[2001], p. 5).

### 4.4 Finsler's rejection of axiomatics and general 1926 inconsistency result

In addition to skepticism toward axiomatics, there was also skepticism toward set theory and its core claim that only sentences that are derivable from axioms (Zermelo Frankel probably) exist. Swiss mathematician Paul Finsler believed that mathematics exists outside of language (formal sentences). Finsler claimed to have shown incompleteness in formal systems before Godel in 1925 and that his proof was superior because it was not tied to Russell's logic as Godel's was. See "A Restoration the failed: Paul Finsler's theory of sets" in Breger[1996], p. 257 for discussion of Finsler's result on undecidability and formal proofs and its history (also Finsler[1996] and Finsler[1969]).

## 5. Turing Machine incorrect model for computation

The central argument for AI is based on the Church Turing thesis. Namely that Turing machines (TM) are universal and anything that involves intelligence can be calculated by TMs. Applying Lighthill's combinatorial explosion arguments, it seems to me that TMs are the wrong model of computation. Instead a different computational model called MRAMS (random access machines with unit multiply and a bounded number of unbounded size memory cells) is a better model of computation (Meyer[2016]). Von Neumann understood the need for random access memory in his design of the von Neumann architecture (pp. 5-6). For MRAM machines deterministic and non deterministic computations are both solvable in polynomial bound time so at least for some problems in the class NP, the combinatorial explosion is mitigated. TM's are universal in the sense that they can compute anything that a von Neumann MRAM machine can calculate. The problem is that Lighthill's combinatorial explosion problem is much worse for TMs than MRAMs because for TMs, problems in NP are probably not computable in polynomial bounded time, but for MRAMs P=NP so there is no advantage to guessing or using heuristics. I claim von Neumann understood that random access and bit select capability that is missing from TMs allows writing programs that avoid combinatorial explosion.

A problem such as asking if two regular expressions are equivalent is outside the class NP. The calculation is inherently exponential on any model that assumes the Church Turing thesis. This suggests that algorithms should be studied as normal data processing because AI's assumption that heuristics and guessing will somehow improve algorithms is problematic.

## 6. Chess - elite human players response to chess programs

Although superiority of chess programs over even the best human chess players is cited as evidence that in the future AI robots will be superior in all areas involving intelligence, in fact the best human chess players now utilize chess position evaluation programs in their preparation so that they are able to defeat even the best chess programs. Although chess seems at first glance to be a simple combinatorial evaluation game, in fact it is one of the best example of human strategic and intuitive thinking that are beyond exhaustive search programming. At best humans can code their (changing) knowledge into computer programs.

AI continues to claim that chess computers are able to defeat the best human players. Reference is usually made to the loss of a match by the at the time world champion Gary Kasperov. In fact Kasperov claims that the IBM team of programmers and chess masters cheated in particular they spied on his team's preparations and changed the Deep Blue Chess program before the file game to execute a line of play its grand masters had developed from analyzing Kasperov's expected line of play (Kasperov[2017], pp. 216-217, for general context 161-220). This shows a common pattern exhibited by much supposed AI success in intelligence requiring activities. Human ingenuity and constant program changes are used to create the illusion of intelligent programs in order to avoid facing the reality that AI is a degenerating research program. Before discussing Kasperov's explanation of his loss, evidence from chess columnists and grand master adaptation to chess programs is discussed.

The world's best chess players are responding in interesting ways to chess computer programs. This corroborates Lighthill's claims that even in a formal sentenced based toy world, combinatorial explosion limits problem solving ability of algorithms. Study of chess playing programs and evaluation of their efficacy show the problems with recent claims of AI successes in general. AI has mostly abandoned discussing chess as a success of AI and now touts other areas where there is less combinatorial explosion. Combinatorial explosion in general and why programs to play the game of 'go' are more successful against humans is discussed in the next section.

Since the 1997 defeat of then world champion Gary Kasperov by the Deep Blue chess program, the world's best chess players have adjusted to computer chess programs. In the December 31, 2016 Financial Times newspaper chess column, Leonard Barton referring to US champion Fabiano Caruana writes: "The US champion and world No. 2 unleashed a brilliant opening novelty, which incidentally showed the limitations of the most powerful computers" (Barton[2016]).

In the October 14, 2017 Financial Times weekend edition, Barton discusses newer responses of the best chess players to computer chess programs. The best human chess players are changing to what are seemingly inferior opening such as Magnus Carlsen's A3 (left rook pawn advances one square) because "Grand masters are turning to the byways of opening theory as powerful programs analyze main lines to a depth unimaginable before the age of computers." Computer chess program "intelligence" is not way beyond human skill, but computers are a tool that allows large improvement in human ability to analyze chess positions. This is similar to microscopes as tools that allow understanding biological cells in previously impossible ways.

Also, US champion Fabio Caruana is still in the for front of using computers as a tool to analyze positions. He found a variation on a well established opening. "Caruana found a nuance at move 19 which was so strong that he had a won game while still in his prep." Barton sums up the reaction to computers as "Carlson's message is clear. Offbeat openings can save a lot of wasted preparation."

It has taken two decades and Caruana was only five years old when Kasperov lost to Deep Blue, but it appears computer algorithms will encounter combinatorial explosion problems so that more and more of the best players will be able to defeat computers. Chess has received new funding allowing chess position evaluation programs to be developed to the specifications of the best grand masters.

IBM dismantled Deep Blue after the 1997 match. There have not been official matches between computers and grand masters since then. One response to computer chess programs is popularity of a chess variant developed by Bobby Fischer called chess960. The 960 comes from the 960 different initial chess piece positions possible when the back rows are randomized, but in such a way that bishops are still opposite color and castling is possible. The new popularity of this chess variant is in response to computer applications that allow computer programs and less skilled players to memorize games. Chess960 eliminates the ability to use large sized files of analyzed game openings. Kasperov[2017] discusses the analyzed position data base problem (pp. 206-208).

In May 2017, Garry Kasperov published a book on his 1997 match against Deep Blue "Deep Thinking: Where Machine Intelligence Ends and Human Creativity Begins" (Kasperov[2017]). Kasperov blames the Deep Blue team cheating (psychological harassment) for his loss. It was Kasperov versus a group of chess masters with access to a very good chess position analysis tools versus Kasparov alone in Kasperov's view (pp. 215-220). This illustrates a common pattern in computer applications programming. Injection of human knowledge by means of writing computer code is an integral part of machine learning. People will even continually change programs in response to threats of falsification or defeat even to the extent of adding commands to allow easy algorithm overriding.

Possibly more interesting is how the injection of human knowledge shows problems with AI scientific methodology. There is an unwillingness to openly specify potential falsifiers of AI as required by Popperian falsification. It is also difficult to criticize AI because of the uneven financial incentives against criticizing it. The Kasperov Deep Blue match is an example. Kasperov made more money by losing rather than by winning. From Kasperov's viewpoint he could win and go back to collecting meager chess tournament prize money or lose and collect appearance fees not just from chess appearances but also from appearances as a marketing representative. It is even possible that Kasperov saw the less than honest IBM team behavior coming because he negotiated a $700 thousand to winner and $400 thousand to loser nearly equal prize package. AI claims of success involving human competition with computers usually follow this pattern. At a minimum, AI tests of this type need to use double blind protocols. A better method for determining if computers can defeat the best human players would be to use double blind tournaments where opponents may be humans or computers and participants and officials were not allowed to know who was who and rebooting programs during matches would not be allowed.

Finally, progress in chess playing computer programs shows that chess programs are normal data processing applications in the Lighthill sense in which human knowledge of chess can be expressed and amplified by injecting it into computers by writing programs.

## 7. Where combinatorial explosion does not apply

Each application programming problem is different. AI has traditionally used chess as a paradigm for computer programs that require intelligence. Since it has turned out that chess requires long term strategic thinking and since it appears chess programs can not defeat the best human players, 'go' is now the game whose computer programs are claimed to really display intelligence. Computer programs that play the game of 'go' are able to defeat the best human players because primitive operations are few and simple. After discussing 'go', this section considers other types of possible combinatorial explosion.

### 7.1 The game of 'go'

The 'go' game has only two types of pieces called stones. One color for each player. A move consists of placing one stone of one's own color on an empty intersection on the board. The reason why combinatorial explosion search limitations do not apply to 'go' is best understood by studying why Braid group cryptography is not used. Braid group public key cryptography seems like a good secrecy method because braid groups lack patterns and have a very large number of states seemingly making exhaustive search for plain text infeasible. Braid group cryptography is a type of public key cryptography that works by publishing a Braid group and requiring finding the inverse of the group to break the code (Garber[2008] for a survey). The primitive group operation for Braid groups is juxtaposition (called concatenation). This substring rearrangement operation is so simple that hardware and parallel exhaustive search is possible. Von Neumann style computers can execute this operation so fast that the size of Braid group keys needs to be infeasibly large.

The situation with 'go' is the same. Von Neumann style computers can exhaustively search for stone arrangements because of the simplicity of a move. The 19 by 19 grid is too small allowing exhaustive search while the board size is large enough so that people are unable to perceive patterns.

There are many ways of writing 'go' programs to perform exhaustive search. The current favorite search method uses programs that optimize search by competing against themselves. AI then makes the claim that computers are teaching each other how to be intelligent so they will soon make humans obsolete. People are supposed to worry about what happens when they are intellectually inferior. There is a problem with this line of reasoning because exhaustive search used in the programs that play themselves to 'learn' is just optimization of evaluation function parameters. There are many ways to optimize parameters such as human choice, dynamic programming or the now popular deep learning game playing.

One approach developed by Popperians in falsifying pseudo scientific theories involves studying not the sophisticated arguments defending a theory, but rather to consider the popular expositions of it (Cioffi[1986]). There is an interesting press release (Lathrop[2016]) with title "Software That Beat World's Best Human 'go' Player Has Roots in UMass Amherst Computer Science" that I think explains what machine learning really is. The release begins by stating "'go' is so complex that experts in AI once believed it would be decades, if ever, before a computer could beat the best human player." The release describes how the AlphaGo program developed by Google DeepMind defeated the world 'go' champion. It continues quoting professor Sridhar Mahadevan.

*This is a huge breakthrough for AI and machine learning. AlphaGo was not programmed explicitly to play 'go', but learned to play using an approach called reinforcement learning, or RL (Sutton[2018] for an RL textbook).*

Mahadevan continues.

*As Mahadevan explains, the standard approach to machine learning, also known as data mining, applied to strategy games is to learn from a large database of moves played by human experts. It works by teaching a machine to quickly search its database for the best move it can make for every position on the board. "But as the game gets more complicated, it doesn't work," he says. "It has a ceiling; the calculations get so numerous that this 'human supervised' approach to machine learning has never gotten a machine to play games at the championship level. It turns out that in order to get to the highest level, the machine crucially has to also play against itself and learn how to win by trial and error, or reinforcement learning.*

The release continues quoting professor Andy Barto.

*Something called the search space for a game, Barto explains, contains all possible games that could be played, and for 'go' it is astronomically huge, many times larger than for chess, for example. There is a "combinatorial explosion," he notes, because games can take a lot of plays and so many moves are possible at each play.*

*"But interestingly that's not the main problem," he says. "The real problem, the experts tell me, is coming up with a good evaluation function. That is a way of looking at a board configuration and assigning it a value for how good it would be in terms of winning the game. It's a prediction of your chances of winning from that configuration, and it has to be quickly computable."*

*"Short of planning moves by doing a lot of look ahead searches, you need a shortcut to predict, a very quick assessment without doing all that look ahead. That's what AlphaGo has come up with and what many people didn't think was possible. It turns out that a good evaluation function can circumvent the combinatorial explosion," Barto adds.*

A 'good evaluation function' allows 'go' to avoid combinatorial explosion. The claim that the reinforcement learning search method is intelligence and is a general method that does not require problem specific knowledge such as good data structures for storing game states and a good evaluation functions is not correct. I view the innovation as developing the evaluation function and determining parameters to optimize. I wonder if anyone is looking for better search algorithms for finding good parameters.

## 7.2 Searching for molecules with medicinal properties

Application programs for searching chemical molecule data bases for particular properties also avoids combinatorial explosion. The area is called Cheminformatics. There are seemingly a very large number of feasible combinations of atoms in biological molecules. However, there are so many chemical and physical constraints that the search space is not infeasibly large. The constraints probably go back to the Pauli exclusion principle. I think the most important innovation in this area was developing good data structures for storing molecules in three dimensions (Jurs[1996] for an advanced text book).

## 7.3 Climate models - Dyson's criticism based on combinatorial explosion

Physicist Freeman Dyson criticizes climate models by arguing that the continuous physical space being modeled has too many states for models to be valid (Dyson[2009]). This is an area for which there is controversy whether climate models run into a combinatorial explosion or not. This is a different non digital type of combinatorial explosion. Dyson begins.

*That's how I got interested. There was an outfit called the Institute for Energy Analysis at Oak Ridge. I visited Oak Ridge many times, and worked with those people, and I thought they were excellent. And the beauty of it was that it was multi-disciplinary. There were experts not just on hydrodynamics of the atmosphere, which of course is important, but also experts on vegetation, on soil, on trees, and so it was sort of half biological and half physics. And I felt that was a very good balance.*

Dyson describes physical model combinatorial explosion this way.

*What's wrong with the models. I mean, I haven't examined them in detail, (but) I know roughly what's in them. And the basic problem is that in the case of climate, very small structures, like clouds, dominate. And you cannot model them in any realistic way. They are far too small and too diverse.*

*So they say, 'We represent cloudiness by a parameter,' but I call it a fudge factor. So then you have a formula, which tells you if you have so much cloudiness and so much humidity, and so much temperature, and so much pressure, what will be the result... But if you are using it for a different climate, when you have twice as much carbon dioxide, there is no guarantee that that's right. There is no way to test it.*

## 8. Why it is difficult to criticize AI

Solving problems and using computers to accomplish tasks is an important tool for modern society. The problem is that AI has a fantasy view that there is just one problem that once computers solve, there will be no need for human problem solving or human knowledge. Nils Nilsson's survey book *The Quest for Artificial Intelligence - A history of Ideas and Achievements* (Nilsson[2010]) exhibits a pattern that Popperian's have identified as a common trait of pseudo sciences that make them hard to test let alone falsify. Frank Cioffi in has falsification of Freudian psychoanalysis describes the problem as concrete claims in the absence of criticism but general discussion listing various aspects of a theory in the face of criticism (Cioffi[1985]). This mention of combinatorial explosion in Nilsson's book shows the lack of awareness of impossibility of AI from combinatorial explosion.

*Although these results from complexity theory did constitute one of the âspeed bumpsâ in AI's rapid forward progress, AI researchers quickly recovered and found various ways around the combinatorial explosion problem. They pointed out, for example, that complexity results were based on worst-case performance, and solutions might often be found faster than in the worst case* (p. 404).

Part of the problem with AI's lack of concrete testable claims is the result of CS tendency to view all problems as just one problem. It is possible to abstract out properties that are common to many problems such as Richard Karp's reduction of many optimization problems to the satisfiability problem from set theory (Karp[1972]). The problem with Karp's reduction is that there is not just one problem because it is also possible to abstract out problem specific properties. There is no way to decide which abstraction should be chosen. There is also evidence that there is even no way to decide between models of algorithm complexity such as Turing machine unbounded memory size NP completeness versus MRAM data indexing capability (Meyer[2010]). In my view application programming is a better research program than attempting to find artificial intelligence. Application programming possibly using Naur's paradigm benefits from proliferation of different applications and theories as advocated by Popperians.

## 9. Conclusion - suggestion to replace AI with Naur's Dataology

A problem with this paper is that people trained to perform advanced computational research before the mid 1970s (primarily physicists) are unable to imagine AI as having any content, but people trained after CS became formalized as object oriented programming, computer programs verified by correctness proofs and axiomatized proofs of algorithm efficiency can not imagine anything but computation as artificial intelligence expressed as formalized logic. Computation researchers trained after the 1970s are unable to imagine alternatives to the AI dogmas. One suggestion is to adopt the ideas of Danish computer scientist and Turing Award winner Peter Naur. Naur was trained as a physicist before pioneering development of computer programming languages. Naur argued that computation should be studied as dataology. Dataology is a theory neutral term for studying data. Naur wrote "mental life during the twentieth century has become entirely misguided into an ideological position such that only discussions that adopt the computer inspired form" are accepted. (Naur[2007], p. 87).

In the 1990s, Naur realized that CS had become too much formal mathematics separated from reality. Naur advocated the importance of programmer specific program development that does not use preconceptions. I would put it: computation allows people to express their ideas by writing computer programs.

The clearest explanation for Naur's paradigm is expressed in the book *Conversations - Pluralism in*

*Software Engineering* (Naur[2011]). This books amplifies the program development method Naur described in his 2005 Turing Award lecture (Naur[2007]). In Naur[2011] page 30, the interviewer asks "... you basically say that there are no foundations, there is no such thing as computer science, and we must not formalize for the sake of formalization alone." Naur answers, "I am not sure I see it this way. I see these techniques as tools which are applicable in some cases, but which definitely are not basic in any sense." Naur continues (p. 44) "The programmer has to realize what these alternatives are and then choose the one that suits his understanding best. This has nothing to do with formal proofs." Dataology without preconceptions and predictions of imminent replacement of human intelligence by robots would improve the scientific study of computation. People not convinced by my arguments but who believe in the Popperian scientific method should attempt to falsify Naur's dataology research programme.